\pdfoutput=1
\documentclass[11pt]{article}

\usepackage[]{acl}

\usepackage{times}
\usepackage{latexsym}

\usepackage[T1]{fontenc}
\usepackage[utf8]{inputenc}

\usepackage{microtype}

\usepackage{bm}
\usepackage{amsmath}
\usepackage{url}
\usepackage{graphicx}
\usepackage{float} 
\usepackage{subfigure}
\usepackage{multirow}
\usepackage{makecell}
\usepackage{xpinyin}
\usepackage{enumerate}

\usepackage{xcolor}
\usepackage{soul}
\newcolumntype{L}[1]{>{\raggedright\let\newline\\\arraybackslash\hspace{0pt}}m{#1}}
\setul{2pt}{2pt}
\usepackage{booktabs}
\usepackage{amsmath,amsfonts}%conflict with \usepackage{bbold}
\usepackage[utf8]{inputenc}
\usepackage{cleveref}

\crefname{section}{§}{§§}
\Crefname{section}{§}{§§}
\usepackage{enumitem}
\setenumerate[1]{itemsep=0pt,partopsep=0pt,parsep=\parskip,topsep=5pt}
\setitemize[1]{itemsep=0pt,partopsep=0pt,parsep=\parskip,topsep=5pt}
\setdescription{itemsep=0pt,partopsep=0pt,parsep=\parskip,topsep=5pt}
\usepackage{arydshln}

\usepackage[linesnumbered,ruled,vlined]{algorithm2e}

\title{Scheduled Multi-task Learning for Neural Chat Translation}

\author{
  Yunlong Liang\textsuperscript{1}\thanks{ \ \ Work was done when Yunlong was interning at Pattern Recognition Center, WeChat AI, Tencent Inc, China.}  , 
  Fandong Meng\textsuperscript{2}, 
  \textbf{Jinan Xu}\textsuperscript{1}\thanks{ \ \ Jinan Xu is the corresponding author.}  , 
  \textbf{Yufeng Chen}\textsuperscript{1}
   and \textbf{Jie Zhou}\textsuperscript{2}\\
  \textsuperscript{1}Beijing Key Lab of Traffic Data Analysis and Mining, \\Beijing Jiaotong University, Beijing, China \\
  \textsuperscript{2}Pattern Recognition Center, WeChat AI, Tencent Inc, China \\
  \texttt{\{yunlongliang,jaxu,chenyf\}@bjtu.edu.cn} \\
  \texttt{\{fandongmeng,withtomzhou\}@tencent.com} \\
}

\begin{document}
\begin{CJK}{UTF8}{gkai}
\maketitle
\begin{abstract}
The goal of the Neural Chat Translation (NCT) is to translate conversational text into different languages. Existing methods mainly focus on modeling the bilingual dialogue characteristics (\emph{e.g.}, coherence) to improve chat translation via multi-task learning on small-scale chat translation data. Although the NCT models have achieved impressive success, it is still far from satisfactory due to insufficient chat translation data and simple joint training manners. To address the above issues, we propose a scheduled multi-task learning framework for NCT. Specifically, we devise a three-stage training framework to incorporate the large-scale in-domain chat translation data into training by adding a second pre-training stage between the original pre-training and fine-tuning stages. Further, we investigate where and how to schedule the dialogue-related auxiliary tasks in multiple training stages to effectively enhance the main chat translation task. Extensive experiments on four language directions (English$\leftrightarrow$Chinese and English$\leftrightarrow$German) demonstrate the effectiveness of the proposed approach. Additionally, we have made the large-scale in-domain paired bilingual dialogue dataset publicly available for the research community.\footnote{The code and in-domain data are publicly available at: \url{https://github.com/XL2248/SML}} %\footnote{The data samples are attached as supplementary materials and our code with the data will be publicly available once accepted.}
\end{abstract}

\section{Introduction}
A bilingual conversation involves speakers in two languages (\emph{e.g.}, one speaking in Chinese and another in English), where a chat translator can be applied to help them communicate in their native languages. The chat translator bilaterally converts the language of bilingual conversational text, \emph{e.g.} from Chinese to English and vice versa~\cite{wang-etal-2016-automatic,farajian-etal-2020-findings,liang-etal-2021-modeling,liang2022msctd}. %With the rapid development of globalization, it becomes more and more necessary and has gained widespread applications. 
\textbf{\begin{figure}[t]
    \centering
    \includegraphics[width=0.49\textwidth]{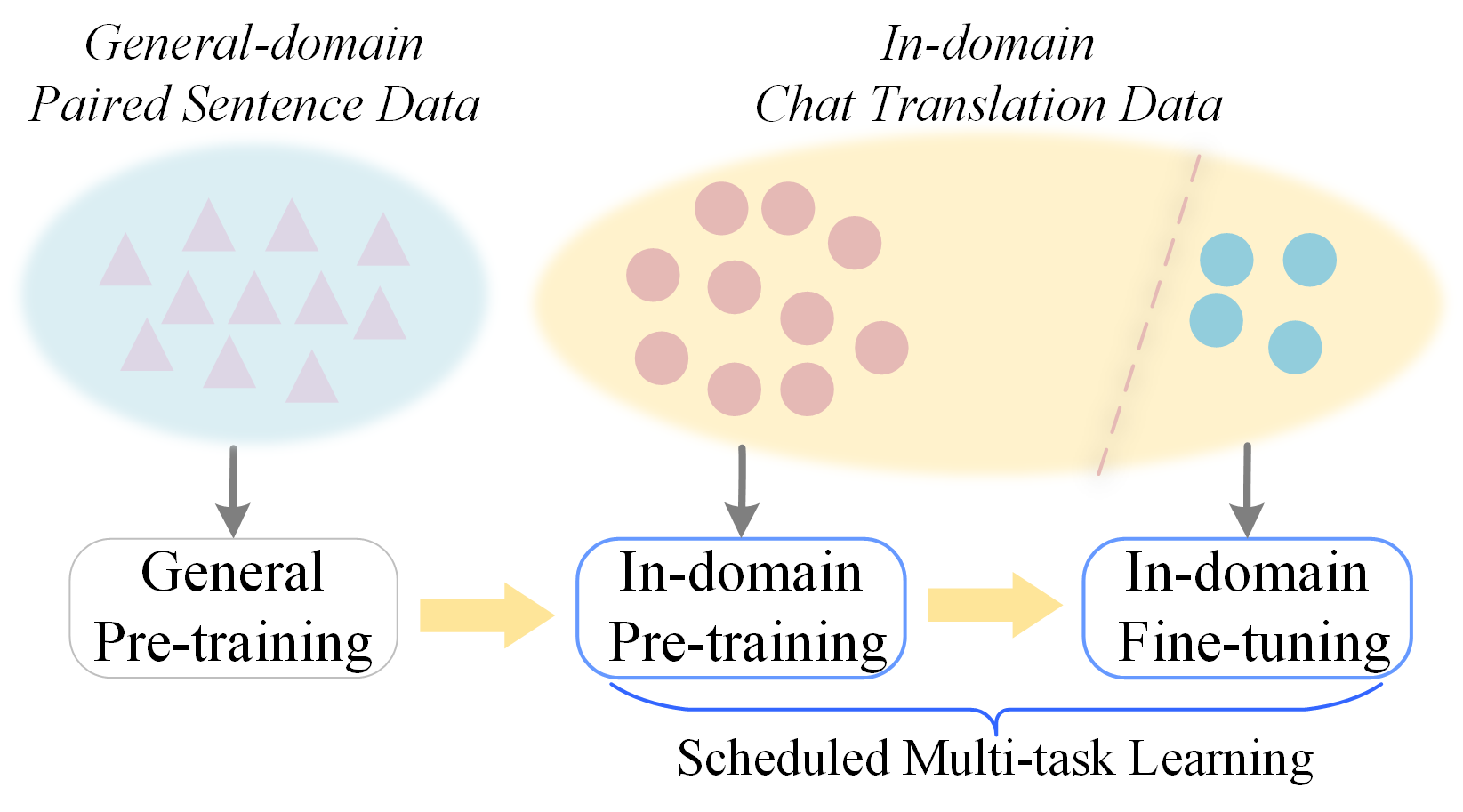}
    \caption{The overall three-stage training framework.
    }
    \label{fig.1}%\vspace{-10pt}
\end{figure}}

Generally, since the bilingual dialogue corpus is scarce, researchers~\cite{bao-EtAl:2020:WMT,wang-EtAl:2020:WMT1,liang-etal-2021-modeling,liang-etal-2021-towards} resort to making use of the large-scale general-domain data through the pre-training-then-fine-tuning paradigm as done in many context-aware neural machine translation models (\citealp{miculicich-etal-2018-document,maruf-haffari-2018-document,tiedemann-scherrer-2017-neural,maruf-etal-2019-selective,voita-etal-2018-context,voita-etal-2019-context,voita-etal-2019-good,yang-etal-2019-Capsule,wang-etal-2019-one,tu-etal-2018-learning,ma-etal-2020-simple}, etc), having made significant progress. However, conventional pre-training on large-scale general-domain data usually learns general language patterns, which is also aimless for capturing the useful dialogue context to chat translation, and fine-tuning usually suffers from insufficient supervised data (about 10k bilingual dialogues). %, preventing general pre-training from effectively capturing these patterns. 
% However, the pre-training  which is insufficient for data-lack fine-tuning to adapt to chat translation task. 
Some studies~\cite{gu-etal-2020-train,gururangan-etal-2020-dont,liu-etal-2021-multi,moghe-hardmeier-bawden:2020:WMT,wang-EtAl:2020:WMT1,ruder2021lmfine-tuning} have shown that learning domain-specific patterns by additional pre-training is beneficial to the models. To this end, we firstly construct the large-scale in-domain chat translation data\footnote{Firstly, to build the data, for English$\leftrightarrow$Chinese (En$\leftrightarrow$Zh), we crawl two consecutive English and Chinese movie subtitles (not aligned). For English$\leftrightarrow$German (En$\leftrightarrow$De), we download two consecutive English and German movie subtitles (not aligned). Then, we use several advanced technologies to align En$\leftrightarrow$Zh and En$\leftrightarrow$De subtitles. Finally, we obtain the paired bilingual dialogue dataset. Please refer to~\autoref{sec:ts} for details.}. And to incorporate it for learning domain-specific patterns, we then propose a three-stage training framework via adding a second pre-training stage between general pre-training and fine-tuning, as shown in~\autoref{fig.1}. 

To further improve the chat translation performance through modeling dialogue characteristics (\emph{e.g.}, coherence), inspired by previous studies~\cite{phang-etal-2020-english,liang-etal-2021-towards,pruksachatkun-etal-2020-intermediate}, we incorporate several dialogue-related auxiliary tasks to our three-stage training framework. Unfortunately, we find that simply introducing all auxiliary tasks in the conventional multi-task learning manner does not obtain significant cumulative benefits as we expect. It indicates that the simple joint training manner may limit the potential of these auxiliary tasks, which inspires us to investigate where and how to make these auxiliary tasks work better for the main NCT task.

To address the above issues, we present a \textbf{S}cheduled \textbf{M}ulti-task \textbf{L}earning framework (SML) for NCT, as shown in~\autoref{fig.1}. Firstly, we propose a three-stage training framework to introduce our constructed in-domain chat translation data for learning domain-specific patterns. Secondly, to make the most of auxiliary tasks for the main NCT task, \textbf{where}: we analyze in which stage these auxiliary tasks work well and find that they are \emph{different strokes for different folks}. Therefore, to fully exert their advantages for enhancing the main NCT task, \textbf{how}: we design a gradient-based strategy to dynamically schedule them at each training step in the last two training stages, which can be seen as a fine-grained joint training manner. In this way, the NCT model is effectively enhanced to capture both domain-specific patterns and dialogue-related characteristics (\emph{e.g.}, coherence) in conversation, which thus can generate better translation results. 

We validate our SML model on two datasets: BMELD~\cite{liang-etal-2021-modeling} (En$\leftrightarrow$Zh) and BConTrasT~\cite{farajian-etal-2020-findings} (En$\leftrightarrow$De). Experimental results demonstrate that the proposed method gains consistent improvements on four translation tasks in terms of both BLEU~\cite{papineni2002bleu} and TER~\cite{snover2006study} scores, showing its superiority and generalizability. Human evaluation also suggests that the SML can produce more coherent and fluent translations than related methods. 

Our contributions are summarized as follows:
\begin{itemize}
% \item To the best of our knowledge, we are the first to incorporate the dialogue coherence and speaker personality into neural chat translation.
\item We propose a scheduled multi-task learning framework with three training stages, where a gradient-based scheduling strategy is designed to fully exert the auxiliary tasks' advantages for the main NCT task, for higher translation quality.

\item Extensive experiments on four chat translation tasks show that our model achieves new state-of-the-art performance and outperforms the existing NCT models by a significant margin.

\item We contribute two large-scale in-domain paired bilingual dialogue corpora (28M for En$\leftrightarrow$Zh and 18M for En$\leftrightarrow$De) to the research community.
\end{itemize}

\section{Background: Conventional Multi-task Learning for NCT}
We introduce the conventional multi-task learning framework~\cite{liang-etal-2021-towards} for NCT, which includes four parts: \emph{problem formalization} (\autoref{sec:pf}), \emph{the NCT model} (\autoref{sec:nct}), \emph{existing three auxiliary tasks} (\autoref{sec:aux}), and \emph{training objective} (\autoref{sec:cto}). 
%%%%%%%%%%%%%%%%%%%%%%%%%%%%%%%%%%%%%%%%%%%%%%%%% DocNMT %%%%%%%%%%%%%%%%%%%%%%%%%%%%%%%%%%%%%%%%%%%%%%%%
\subsection{Problem Formalization}
\label{sec:pf}
We assume in a bilingual conversation the two speakers have given utterances in two languages for $u$ turns, resulting in $X_1, X_2, X_3,..., X_u$ and $Y_1, Y_2, Y_3,...,Y_u$ on the source and target sides, respectively. In these utterances, $X_1, X_3, X_5, X_7,..., X_u$ are originally spoken and $Y_1, Y_3, Y_5, Y_7,..., Y_u$ are the corresponding translations in the target language. On the other side, $Y_2, Y_4, Y_6, Y_8,..., Y_{u-1}$ are originally spoken and $X_2, X_4, X_6, X_8,..., X_{u-1}$ are the translated utterances in the source language. According to different languages, we define the dialogue context of $X_u$ on the source side as $\mathcal{C}_{X_u}$=\{$X_1, X_2, X_3, ..., X_{u-1}\}$ and that of $Y_u$ on the target side as $C_{Y_u}$=\{$Y_1, Y_2, Y_3,..., Y_{u-1}\}$.\footnote{For each of \{$C_{X_u}$, $C_{Y_u}$\}, we append a special token `[CLS]' tag at the head of it and add another token `[SEP]' to delimit its included utterances, as done in~\citet{bert}.}

Then, an NCT model aims to translate $X_u$ to $Y_u$ with dialogue context $\mathcal{C}_{X_u}$ and $C_{Y_u}$.%\footnote{Here, we just take one translation direction (\emph{i.e.}, En$\rightarrow$Zh) as an example, which is similar for other directions.} %Next, we describe the NCT model.
%%%%%%%%%%%%%%%%%%%%%%%%%%%%%%%%%%%%%%%%%%%%%%%%% DocNMT %%%%%%%%%%%%%%%%%%%%%%%%%%%%%%%%%%%%%%%%%%%%%%%%
% \subsection{Multi-task Training Framework.}
\subsection{The NCT Model}
\label{sec:nct}
The NCT model~\cite{ma-etal-2020-simple,liang-etal-2021-towards} utilizes the standard transformer~\cite{vaswani2017attention} architecture with an encoder and a decoder\footnote{Here, we just describe some adaptions to the NCT model, and please refer to~\citet{vaswani2017attention} for more details.}. 

In the encoder, it inputs $[\mathcal{C}_{X_u}$; $X_u]$, where $[;]$ denotes the concatenation operation. The input embedding consists of word embedding $\mathbf{WE}$, position embedding $\mathbf{PE}$, and turn embedding $\mathbf{TE}$:
\begin{equation}\label{input_embed}\nonumber
\setlength{\abovedisplayskip}{5pt}
\setlength{\belowdisplayskip}{5pt}
\mathbf{B}(x_i) = \mathbf{WE}({x_i}) + \mathbf{PE}({x_i}) + \mathbf{TE}({x_i}),
\end{equation}
where $\mathbf{WE}\in\mathbb{R}^{|V|\times{d}}$ and $\mathbf{TE}\in\mathbb{R}^{|T|\times{d}}$.\footnote{$|V|$, $|T|$ and $d$ denote the size of shared vocabulary, maximum dialogue turns, and the hidden size, respectively.} When computation in the encoder, as done in~\citet{ma-etal-2020-simple}, tokens in the context $\mathcal{C}_{X_u}$ can only be attended by those in the utterance $X_u$ at the first encoder layer while the context $\mathcal{C}_{X_u}$ is masked at the other layers.

%%%%%%%%%%%%%%%%%%%%%%%%%%%%%%%%%%%%%%%%%%%%%%%%% Decoder %%%%%%%%%%%%%%%%%%%%%%%%%%%%%%%%%%%%%%%%%%%%%%%%
In the decoder, at each decoding time step $t$, the top-layer ($L$-th) decoder hidden state $\mathbf{h}^{L}_{d,t}$ is fed into the softmax layer to produce the probability distribution of the next target token as:
\begin{equation}
\setlength{\abovedisplayskip}{5pt}
\setlength{\belowdisplayskip}{5pt}
\resizebox{1.05\hsize}{!}{$
\begin{split}
    p(Y_{u,t}|Y_{u,<t}, \mathcal{C}_{X_u}, X_u) &= \mathrm{Softmax}(\mathbf{W}_o\mathbf{h}^{L}_{d,t}+\mathbf{b}_o),\nonumber
\end{split}
$}
\end{equation}
where $Y_{u,<t}$ indicates the previous tokens before the $t$-th time step in $Y_u$, $\mathbf{W}_o$ and $\mathbf{b}_o$ are trainable weights. 

Finally, the loss function is defined as:
\begin{equation}
\setlength{\abovedisplayskip}{5pt}
\setlength{\belowdisplayskip}{5pt}
\begin{split} 
\label{eq:cnmt}
    \mathcal{L}_{\text{NCT}} = -\sum_{t=1}^{|Y_u|}\mathrm{log}(p(Y_{u,t}|Y_{u,<t}, \mathcal{C}_{X_u}, X_u)).%\nonumber
\end{split}
\end{equation}

\subsection{Existing Auxiliary Tasks}
\label{sec:aux}
To generate coherent translation, ~\citet{liang-etal-2021-towards} present Monolingual Response Generation (MRG) task, Cross-lingual Response Generation (XRG) task, and Next Utterance Discrimination (NUD) task during the NCT model training. 

\paragraph{MRG.} This task aims to help the NCT model to produce the utterance $Y_u$ coherent to $\mathcal{C}_{Y_u}$ given the dialogue context $\mathcal{C}_{Y_u}$ in the target language. Specifically, the encoder of the NCT model is used to encode $\mathcal{C}_{Y_u}$, and the decoder is used to generate $Y_u$. Formally, the training objective of the MRG is defined as:
\begin{equation}%\nonumber
\setlength{\abovedisplayskip}{5pt}
\setlength{\belowdisplayskip}{5pt}
% \resizebox{.76\hsize}{!}{$
\begin{split}
    \mathcal{L}_{\text{MRG}}  =-\sum^{|Y_u|}_{t=1}\mathrm{log}(p(Y_{u,t}|Y_{u,<t}, \mathcal{C}_{Y_u})),\\
    {p}(Y_{u,t}|Y_{u,<t}, \mathcal{C}_{Y_u}) = \mathrm{Softmax}(\mathbf{W}_{m}\mathbf{h}^{L}_{d,t}+\mathbf{b}_{m}),\nonumber
\end{split}
% $}
\end{equation}
where $\mathbf{h}^{L}_{d,t}$ is hidden state at the $t$-th decoding step of the $L$-th decoder, $\mathbf{W}_{m}$ and $\mathbf{b}_{m}$ are trainable weights. 

\paragraph{XRG.} 
Similar to MRG, the goal of the XRG is to generate the utterance $Y_u$ coherent to the given dialogue context $\mathcal{C}_{X_u}$ in the source language:
\begin{equation}%\nonumber
\setlength{\abovedisplayskip}{5pt}
\setlength{\belowdisplayskip}{5pt}
% \resizebox{.76\hsize}{!}{$
\begin{split}
    \mathcal{L}_{\text{XRG}}  =-\sum^{|Y_u|}_{t=1}\mathrm{log}(p(Y_{u,t}|Y_{u,<t}, \mathcal{C}_{X_u})),\\
    {p}(Y_{u,t}|Y_{u,<t}, \mathcal{C}_{X_u}) = \mathrm{Softmax}(\mathbf{W}_{c}\mathbf{h}^{L}_{d,t}+\mathbf{b}_{c}), \nonumber
\end{split}
% $}
\end{equation}
where $\mathbf{W}_{c}$ and $\mathbf{b}_{c}$ are trainable parameters. 

% Note that in the above two response generation tasks, we use the same set of NCT model parameters except for the softmax layer (\emph{i.e.}, $\mathbf{W}_{m}$, $\mathbf{b}_{m}$, $\mathbf{W}_{c}$ and $\mathbf{b}_{c}$).

\paragraph{NUD.} 
This task aims to judge whether the translated text is coherent to be the next utterance of the given dialogue context. Specifically, the negative and positive samples are firstly constructed: (1) the negative sample (${Y}_{u^{-}}$, $\mathcal{C}_{Y_{u}}$) with the label $\ell=0$ consists of the dialogue context $\mathcal{C}_{Y_{u}}$ and a randomly selected utterance ${Y}_{u^{-}}$ from the preceding context of ${Y}_{u}$; (2) the positive one (${Y}_{u^{+}}$, $\mathcal{C}_{Y_{u}}$) with the label $\ell=1$ consists of the identical context $\mathcal{C}_{Y_{u}}$ and the target utterance ${Y}_{u}$. Finally, the training loss of the NUD task is written as:
\begin{equation}\nonumber
\setlength{\abovedisplayskip}{5pt}
\setlength{\belowdisplayskip}{5pt}
\begin{split}
    \mathcal{L}_{\text{NUD}}  =&- \mathrm{log}(p(\ell=0|{Y}_{u^{-}}, \mathcal{C}_{Y_{u}}))\\&-\mathrm{log}(p(\ell=1|{Y}_{u^{+}}, \mathcal{C}_{Y_{u}})),\\ 
    {p}(\ell \!=\!1|Y_u, \mathcal{C}_{Y_u})\!=&\ \!\mathrm{Softmax}(\mathbf{W}_{n}[\mathbf{H}_{Y_u}; \mathbf{H}_{\mathcal{C}_{Y_u}}]),
\end{split}
\label{loss_NUD}
\end{equation}
where $\mathbf{H}_{Y_u}$ and $\mathbf{H}_{\mathcal{C}_{Y_u}}$ denote the representation of the utterance ${Y}_u$ and context $\mathcal{C}_{Y_u}$, respectively. Concretely, $\mathbf{H}_{Y_u}=\frac{1}{|Y_u|}\sum_{t=1}^{|Y_u|}\mathbf{h}^{L}_{e,t}$ while $\mathbf{H}_{\mathcal{C}_{Y_u}}$ is the encoder hidden state $\mathbf{h}^{L}_{e,0}$, \emph{i.e.}, the first special token `[CLS]' of $\mathcal{C}_{Y_u}$. The $\mathbf{W}_{n}$ is the trainable weight of the classifier and we also omit the bias term.
\subsection{Training Objective}
\label{sec:cto}
\label{sec:ti}
With the four tasks (NCT, MRG, XRG, and NUD), the training function of the conventional multi-task learning is defined as:
\begin{equation}%\nonumber
\setlength{\abovedisplayskip}{5pt}
\setlength{\belowdisplayskip}{5pt}
\begin{split}
    &\mathcal{L} = \mathcal{L}_{\text{NCT}} + \alpha(\mathcal{L}_{\text{MRG}} + \mathcal{L}_{\text{XRG}} + \mathcal{L}_{\text{NUD}}),
\end{split}\label{loss_all}
\end{equation}
where $\alpha$ is the balancing factor between $\mathcal{L}_{\text{NCT}}$ and the auxiliary losses. 

%%%%%%%%%%%%%%%%%%%%%%%%%%%%%%%%%%%%%%%%%%%%%%%%% Our Method %%%%%%%%%%%%%%%%%%%%%%%%%%%%%%%%%%%%%%%%%%%%%%%%
\section{Scheduled Multi-task Learning for NCT}
In this section, we introduce the proposed \textbf{S}cheduled \textbf{M}ulti-task \textbf{L}earning (SML) framework, including three stages: general pre-training, in-domain pre-training, and in-domain fine-tuning, as shown in~\autoref{fig.1}. Specifically, we firstly elaborate on the process of \emph{in-domain pre-training} (\autoref{sec:ts}) and then present some \emph{findings of conventional multi-task learning} (\autoref{mlf}), which inspire us to investigate the \emph{scheduled multi-task learning} (\autoref{sec:sla}). Finally, we describe the \emph{training and inference} (\autoref{sec:tt}) in detail.
%%%%%%%%%%%%%%%%%%%%%%%%%%%%%%%%%%%%%%%%%%%%%%%%% ChatNMT %%%%%%%%%%%%%%%%%%%%%%%%%%%%%%%%%%%%%%%%%%%%%%%%
\subsection{In-domain Pre-training}
\label{sec:ts}
For the second in-domain pre-training, we firstly build an in-domain paired bilingual dialogue data and then conduct pre-training on it.

To construct the paired bilingual dialogue data, we firstly crawl the in-domain consecutive movie subtitles of En$\leftrightarrow$Zh and download the consecutive movie subtitles of En$\leftrightarrow$De on related websites\footnote{En$\leftrightarrow$Zh: https://www.kexiaoguo.com/ and En$\leftrightarrow$De: https://opus.nlpl.eu/OpenSubtitles.php}. %, about 5,585 movies and 1,628 teleplays. 
Since both bilingual movie subtitles are not strictly aligned, we utilize the Vecalign tool~\cite{thompson-koehn-2019-vecalign}, an accurate sentence alignment algorithm, to align them. Meanwhile, we leverage the LASER toolkit\footnote{https://github.com/facebookresearch/LASER} to obtain the multilingual embedding for better alignment performance. Consequently, we obtain two relatively clean paired movie subtitles. According to the setting of dialogue context length in~\citet{liang-etal-2021-modeling}, we take four consecutive utterances as one dialogue, and then filter out duplicate dialogues. Finally, we attain two in-domain paired bilingual dialogue dataset, the statistics of which are shown in~\autoref{Tbl:data}.
%%%%%%%%%%%%%%%%%%%%%%%%%%%%%% Datasets %%%%%%%%%%%%%%%%%%%%%%%%%%%%%%%%%
\begin{table}[htbp]
\centering
\scalebox{0.88}{
\begin{tabular}{l|rr|r}
\hline
\bf{Datasets}  & \#\bf{Dialogues} & \#\bf{Utterances} & \#\bf{Sentences} \\
\hline
En$\leftrightarrow$Zh    &28,214,769 &28,238,877 &22,244,006 \\
En$\leftrightarrow$De    &18,041,125&18,048,573  &45,541,367 \\
\hline
\end{tabular}}
\caption{Statistics of our constructed chat translation data. The \#\textbf{Sentences} column is the general-domain WMT sentence pairs used in the first pre-training stage.}
\label{Tbl:data}%\vspace{-10pt}
\end{table}

Based on the constructed in-domain bilingual corpus, we continue to pre-train the NCT model after the general pre-training stage, and then go to the in-domain fine-tuning stage, as shown in the In-domain Pre-training\&Fine-tuning parts of~\autoref{fig.1}. 

% Based on the constructed in-domain bilingual corpus, we continue to pre-train the NCT model after the general pre-training stage, and then go to the fine-tuning stage. To further enhance the NCT model, in the last two training process, we investigate on which stages where the auxiliary tasks work well and find that they are \emph{different strokes for different folks}, refer to~\autoref{fig.zx}. Therefore, to make the most of the potential of these auxiliary tasks for enhancing the main NCT task, we propose the scheduled multi-task learning to dynamically schedule them at each training step in the last two training stages, which can be seen as a fine-grained joint training manner.

\subsection{Findings of Conventional Multi-task Learning}
\label{mlf}
According to the finding that multi-task learning can enhance the NCT model~\cite{liang-etal-2021-towards}, in the last two training processes (\emph{i.e.}, the In-domain Pre-training and In-domain Fine-tuning parts of~\autoref{fig.1}), we conduct extensive multi-task learning experiments, aiming to achieve a better NCT model. Firstly, we present one additional auxiliary task,~\emph{i.e.} Cross-lingual NUD (XNUD), given the intuition that more dialogue-related tasks may yield better performance. Then, we conclude some multi-task learning findings that could motivate us to investigate how to use these auxiliary tasks well.

\paragraph{XNUD.}Similar to the NUD task described in~\autoref{sec:aux}, the XNUD aims to distinguish whether the translated text is coherent to be the next utterance of the given cross-lingual dialogue history context. Compared to the NUD task, the different point lies in the cross-lingual dialogue context history, \emph{i.e.}, a negative sample (${Y}_{u^{-}}$, $\mathcal{C}_{X_{u}}$) with the label $\ell=0$ and a positive one (${Y}_{u^{+}}$, $\mathcal{C}_{X_{u}}$) with the label $\ell=1$. Finally, the formal training objective of XNUD is written as follows:
\begin{equation}\nonumber
\setlength{\abovedisplayskip}{5pt}
\setlength{\belowdisplayskip}{5pt}
\begin{split}
    \mathcal{L}_{\text{XNUD}}  =&- \mathrm{log}(p(\ell=0|{Y}_{u^{-}}, \mathcal{C}_{X_{u}}))\\
    &-\mathrm{log}(p(\ell=1|{Y}_{u^{+}}, \mathcal{C}_{X_{u}})),\\ 
    {p}(\ell \!=\!1|Y_u, \mathcal{C}_{X_u})\!=&\ \!\mathrm{Softmax}(\mathbf{W}_{x}[\mathbf{H}_{Y_u}; \mathbf{H}_{\mathcal{C}_{X_u}}]),
\end{split}
\label{loss_XNUD}
\end{equation}
where $\mathbf{H}_{\mathcal{C}_{X_u}}$ denotes the representation of $\mathcal{C}_{Y_u}$, which is calculated as same as $\mathbf{H}_{\mathcal{C}_{Y_u}}$ in NUD. $\mathbf{W}_{x}$ is the trainable weights of the XNUD classifier and the bias term is omitted for simplicity.
\textbf{\begin{figure}[t]
    \centering
    \includegraphics[width=0.49\textwidth]{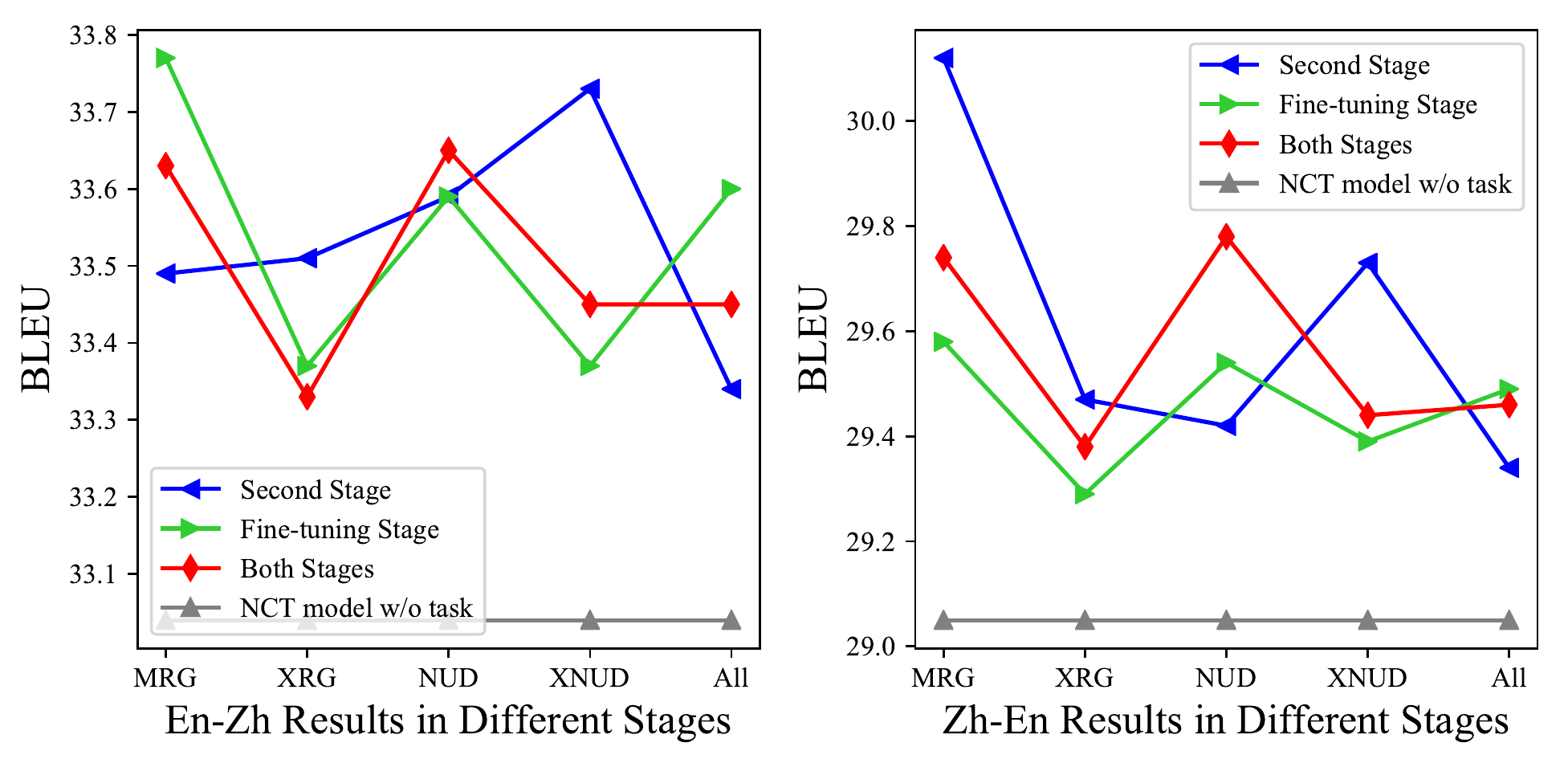}
    \caption{The effect of each task on validation sets in different training stages, under transformer \emph{Base} setting, where ``All'' denotes all four auxiliary tasks. We find that each auxiliary task performs well on the second stage while XRG and XNUD tasks perform relatively poorly in the fine-tuning stage. Further, we observe that all auxiliary tasks in a conventional multi-task learning manner do not obtain significant cumulative benefits. That is, the auxiliary tasks are \emph{different strokes for different folks}.
    }
    \label{fig.zx}%\vspace{-10pt}
\end{figure}}

\paragraph{Findings.} Based on four auxiliary tasks (MRG, XRG, NUD, and XNUD), we investigate in which stage in~\autoref{fig.1} the auxiliary tasks work well in a conventional multi-task learning manner\footnote{Note that, in the last two in-domain stages, we use the conventional multi-task learning to pre-train and fine-tune models rather than the scheduled multi-task learning.} and the following is what we find from~\autoref{fig.zx}: 
\begin{itemize}%[itemindent=1em]
% \item Training with each auxiliary task separately performs better than the NCT model without using any auxiliary task in both the second pre-training and the final fine-tuning stages;
\item Each auxiliary task can always bring improvement compared with the NCT model \emph{w/o} task;
\item By contrast, XRG and XNUD tasks perform relatively poorly in the final fine-tuning stage than MRG and NUD tasks;
% \item Some tasks used only in one stage (\emph{e.g.}, XRG and XNUD in the second stage) perform better than being used in both stages, revealing that throughout using all auxiliary tasks in a simple manner not always performs best; (one best setting seems that all tasks are used in the second stage while only MRG and NUD tasks are used in the final fine-tuning stage.)
% \item Some tasks used only in one stage (\emph{e.g.}, XRG and XNUD in the second stage) perform better than being used in both stages, revealing that all auxiliary tasks used in a simple way not always performs best; (one best setting seems that all tasks are used in the second stage while only MRG and NUD tasks are used in the final fine-tuning stage.)
\item Some tasks used only in one stage (\emph{e.g.}, XRG and XNUD in the second stage) perform better than being used in both stages, revealing that different auxiliary tasks may prefer different stages to exert their advantages; (one best setting seems that all tasks are used in the second stage while only MRG and NUD tasks are used in the final fine-tuning stage.)
\item Using all auxiliary tasks in a conventional multi-task learning manner does not obtain significant cumulative benefits. 
\end{itemize}
Given the above findings, we wonder whether there exists a strategy to dynamically schedule them to exert their potential for the main NCT task.

\subsection{Scheduled Multi-task Learning}
\label{sec:sla}
Inspired by~\citet{NEURIPS2020_3fe78a8a}, we design a gradient-based scheduled multi-task learning algorithm to dynamically schedule all auxiliary tasks at each training step, as shown in Algorithm 1. Specifically, at each training step (\textbf{line 1}), for each task we firstly compute its gradient to model parameters $\theta$ (\textbf{lines 2$\sim$4}, and we denote the gradient of the main NCT task as $\textbf{g}_{nct}$). Then, we obtain the projection of the gradient $\textbf{g}_{k}$ of each auxiliary task $k$ onto $\textbf{g}_{nct}$ (\textbf{line 5}), as shown in~\autoref{fig:gradient}. Finally, we utilize the sum of $\textbf{g}_{nct}$ and all projection (\emph{i.e.}, the blue arrows part, as shown in~\autoref{fig:gradient}) of auxiliary tasks to update model parameters. 

The core ideas behind the gradient-based SML algorithm are: (1) when the cosine similarity between $\textbf{g}_{k}$ and $\textbf{g}_{nct}$ is positive, \emph{i.e.}, the gradient projection $\textbf{g}_k^\prime$ is in the same gradient descent direction with the main NCT task, \emph{i.e.},~\autoref{fig:gradient} (a), which could help the NCT model achieve optimal solution; (2) when the cosine similarity between $\textbf{g}_{k}$ and $\textbf{g}_{nct}$ is negative, \emph{i.e.},~\autoref{fig:gradient} (b), which can avoid the model being optimized too fast and overfitted. Therefore, we also keep the inverse gradient to prevent the NCT model from overfitting as a regularizer. In this way, such auxiliary task joins in training at each step with the NCT task when its gradient projection is in line with $\textbf{g}_{nct}$, which acted as a fine-grained joint training manner.  %we apply the observed best combination of auxiliary tasks in the NCT model and achieve better results. This is laborious
\begin{algorithm}[!t]
    \caption{Gradient-based SML}
			\SetKwData{Index}{Index}
			\SetKwInput{kwInit}{Init}
			\SetKwInput{kwOutput}{Return}
			\SetKwInput{kwInput}{Require}
			\label{alg}
    {
		\kwInput{Model parameters $\theta$, Balancing factor $\alpha$, MaxTrainStep $T$, NCT task, Auxiliary tasks set $\mathcal{T}=\{\text{MRG}, \text{XRG}, \text{NUD}, \text{XNUD}\}$.\\}
        \kwInit{$\theta$, $t=0$}
        \For{$t$ $<$ $T$}{
            $\textbf{g}_{nct}$ $\gets$ $\nabla_\theta$
            $\mathcal{L}_{\text{NCT}}(\theta)$ \\
            \For{$k$ in $\mathcal{T}$}{
            $\textbf{g}_{k}$ $\gets$ $\nabla_\theta$ $\mathcal{L}_{k}(\theta)$\\
            Set $\textbf{g}_k^\prime = \displaystyle \frac{\textbf{g}_k \cdot \textbf{g}_{nct}}{\|\textbf{g}_{nct}\|^2} \textbf{g}_{nct}$
            }
        }
        \kwOutput{Update $\Delta \theta = \textbf{g}_{nct} + \alpha \sum_k \textbf{g}_k^\prime$}
    }
\end{algorithm}

\subsection{Training and Inference}
\label{sec:tt}
Our training process includes three stages: the first pre-training stage on the general-domain sentence pairs ($X$, $Y$):
\begin{equation}%\nonumber
\setlength{\abovedisplayskip}{5pt}
\setlength{\belowdisplayskip}{5pt}
\label{eq:nmt}
    \mathcal{L}_{\text{Sent-NMT}} = -\sum_{t=1}^{|Y|}\mathrm{log}(p(y_t|X, y_{<t})),
\end{equation}
the second in-domain pre-training stage, and the final in-domain  fine-tuning stage on the chat translation data:
\begin{equation}%\nonumber
\setlength{\abovedisplayskip}{5pt}
\setlength{\belowdisplayskip}{5pt}
\begin{split}
    &\mathcal{J} = \mathcal{L}_{\text{NCT}} + \alpha \sum_k^{\mathcal{T}}\mathcal{L}_{k},
\end{split}\label{loss_new}
\end{equation}
where $\mathcal{T}$ is the auxiliary tasks set and we keep the balancing hyper-parameter $\alpha$. Although the form of $\mathcal{L}_{k}$ is the same with \autoref{loss_all}, the gradient that participates in updating model parameters is different where it depends on the gradient descent direction of the NCT task in \autoref{loss_new}.

At inference, all auxiliary tasks are not participated in and only the NCT model after scheduled multi-task fine-tuning is applied to chat translation.

\textbf{\begin{figure}[!t]
    \centering
    \includegraphics[width=0.49\textwidth]{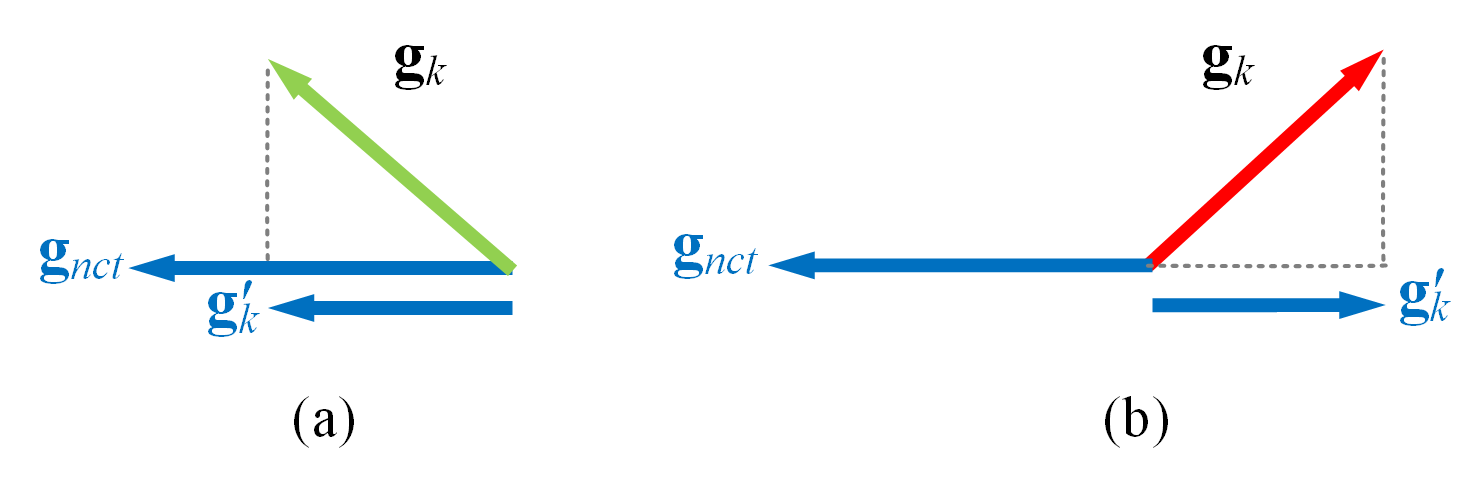}
    \caption{Gradient projection example. 
    }
    \label{fig:gradient}%\vspace{-10pt}
\end{figure}}
%%%%%%%%%%%%%%%%%%%%%%%%%%%%%%%%%%%%%%%%%%%%%%%%% Main results %%%%%%%%%%%%%%%%%%%%%%%%%%%%%%%%%%%%%%%%%%%%%%%%
\begin{table*}[ht!]
\centering
% \scalebox{0.9}{
\begin{tabular}{llllllllll}
\toprule
&\multirow{2}{*}{\textbf{Models}} &  \multicolumn{2}{c}{\textbf{En$\rightarrow$Zh}}  &  \multicolumn{2}{c}{\textbf{Zh$\rightarrow$En}} & \multicolumn{2}{c}{\textbf{En$\rightarrow$De}} & \multicolumn{2}{c}{\textbf{De$\rightarrow$En}}\\
\cmidrule(lr){3-4} \cmidrule(lr){5-6} \cmidrule(lr){7-8} \cmidrule(lr){9-10}
&  & BLEU$\uparrow$ & TER$\downarrow$& BLEU$\uparrow$ & TER$\downarrow$& BLEU$\uparrow$ & TER$\downarrow$& BLEU$\uparrow$ & TER$\downarrow$ \\
\midrule
\multirow{8}{*}{\emph{Base}} 
&{Trans. \emph{w/o} FT}     &21.40     &72.4 & 18.52    & 59.1  & 40.02     & 42.5   & 48.38     & 33.4     \\
&{Trans.}   &25.22 &62.8 & 21.59 &56.7 & 58.43 & 26.7   & {59.57}& 26.2 \\\cdashline{2-10}[4pt/2pt]
&{Dia-Trans.}  &24.96     &63.7 & 20.49   & 60.1  &58.33     &26.8   &59.09    &26.2    \\
&{Gate-Trans.}    &{25.34}    &{62.5} &21.03   &56.9   &{58.48}     &{26.6}   &59.53    &26.1 \\
&{NCT}  & 24.76     &63.4  & 20.61   & 59.8  & 58.15     & 27.1   &59.46    &{25.7}   \\ 
&{CPCC}  &{27.55}      &{60.1}   & \underline{22.50}    & \underline{55.7}   & \underline{60.13} & \underline{25.4}  &\underline{61.05}    &\underline{24.9} \\
&{CSA-NCT}   &\underline{27.77}  &\underline{60.0}   & {22.36}    & {55.9}   & {59.50}   & {25.7}   &{60.65}    &{25.4}\\%\cdashline{2-10}[4pt/2pt]
&{SML (Ours)}   &\textbf{32.25}$^{\dagger\dagger}$     &\textbf{55.1}$^{\dagger\dagger}$  & \textbf{26.42}$^{\dagger\dagger}$   & \textbf{51.4}$^{\dagger\dagger}$   & \textbf{60.65}$^{\dagger}$  & \textbf{25.3}  &\textbf{61.78}$^{\dagger\dagger}$   &\textbf{24.6}$^{\dagger}$  \\
\midrule
\multirow{8}{*}{\emph{Big}}
&{Trans. \emph{w/o} FT}    &22.81  &69.6     &19.58    &57.7  &40.53     &42.2   &49.90 &33.3   \\
&{Trans.}   &26.95 &60.7 &22.15 &56.1 &{59.01} &{26.0} & 59.98  &25.9   \\\cdashline{2-10}[4pt/2pt]
&{Dia-Trans.}  &26.72  &62.4    &21.09    &58.1  &58.68    &26.8  &59.63 &26.0   \\ 
&{Gate-Trans.}    &{27.13}    &{60.3} &{22.26}   &{55.8}   &58.94     &26.2   &{60.08}    &25.5 \\ 
&{NCT}  &26.45  &62.6   &21.38    &57.7  &58.61   &26.5 &59.98 &{25.4}   \\ 
&{CPCC}       &\underline{28.98}  &{59.0} &{22.98}   &\underline{54.6}  &{60.23}  &{25.6}  &\underline{61.45}  &\underline{24.8}\\
&{CSA-NCT}      &{28.86}  &\underline{58.7}  &\underline{23.69}  &{54.7} &\underline{60.64}  &\underline{25.3}  &{61.21}  &{24.9}\\
&{SML (Ours)}   &\textbf{32.87}$^{\dagger\dagger}$  &\textbf{54.4}$^{\dagger\dagger}$  &\textbf{27.58}$^{\dagger\dagger}$  &\textbf{50.6}$^{\dagger\dagger}$ &\textbf{61.16}$^{\dagger}$  &\textbf{25.0}$^{\dagger}$ &\textbf{62.17}$^{\dagger\dagger}$  &\textbf{24.4}$^{\dagger}$ \\ 
\bottomrule
\end{tabular}%}
\caption{Test results on BMELD (En$\leftrightarrow$Zh) and BConTrasT (En$\leftrightarrow$De) in terms of BLEU (\%) and TER (\%). ``$^{\dagger}$'' and ``$^{\dagger\dagger}$'' denote that statistically significant better than the best result of the contrast NMT models with t-test {\em p} \textless \ 0.05 and {\em p} \textless \ 0.01 hereinafter, respectively. The best and second best results are \textbf{bold} and \underline{underlined}, respectively. The results of contrast models are from~\citet{liang-etal-2021-modeling,liang-etal-2021-towards}. Strictly speaking, it is unfair to directly compare with them since we use additional data. Therefore, we conduct further experiments in~\autoref{tbl:w_f} for fair comparison.} 
\label{tbl:main_res}%\vspace{-10pt}
\end{table*}

\section{Experiments}
\subsection{Datasets and Metrics}
\label{sect:data}

\paragraph{Datasets.}\quad The training dataset used in our SML framework includes three parts: (1) a large-scale sentence-level NMT corpus (WMT20\footnote{http://www.statmt.org/wmt20/translation-task.html}), used to pre-train the model; (2) our constructed in-domain chat translation corpus, used to further pre-train the model; (3) the target chat translation corpus: BMELD~\cite{liang-etal-2021-modeling} and BConTrasT~\cite{farajian-etal-2020-findings}. The target dataset details (\emph{e.g.}, splits of training, validation or test sets) are listed in~\autoref{Appendix_dataset}. 

%%%%%%%%%%%%%%%%%%%%%%%%%%%%%%%%%%%%%%%%%%%%%%%%% Metrics %%%%%%%%%%%%%%%%%%%%%%%%%%%%%%%%%%%%%%%%%%%%%%%%
\paragraph{Metrics.}\quad
For a fair comparison, we following~\citet{liang-etal-2021-towards} and adopt SacreBLEU\footnote{BLEU+case.mixed+numrefs.1+smooth.exp+tok.13a+\\version.1.4.13}~\cite{post-2018-call} and TER~\cite{snover2006study} with the statistical significance test~\cite{koehn-2004-statistical}. Specifically, we report character-level BLEU for En$\rightarrow$Zh, case-insensitive BLEU score for Zh$\rightarrow$En, and case-sensitive BLEU score likewise for En$\leftrightarrow$De.

%%%%%%%%%%%%%%%%%%%%%%%%%%%%%%%%%%%%%%%%%%%%%%%%% Implementation %%%%%%%%%%%%%%%%%%%%%%%%%%%%%%%%%%%%%%%%%%%%%%%%
\subsection{Implementation Details}
In our experiments, we mainly utilize the settings of standard \emph{Transformer-Base} and \emph{Transformer-Big} in~\citet{vaswani2017attention}. Generally, we utilize the settings in~\citet{liang-etal-2021-towards} for fair comparison. For more details, please refer to~\autoref{ID}. We investigate the effect of the XNUD task in~\autoref{XNUD}, where the new XNUD performs well based on existing auxiliary tasks.

\subsection{Comparison Models}
\label{ssec:layout}
%%%%%%%%%%%%%%%%%%%%%%%%%%%%%%%%%%%%%%%%%%%%%%%%% Comparison Models %%%%%%%%%%%%%%%%%%%%%%%%%%%%%%%%%%%%%%%%%%%%%%%%
\paragraph{Sentence-level NMT Systems.}

{Trans. \emph{w/o} FT} and {Trans.}~\cite{vaswani2017attention}: both are the de-facto transformer-based NMT models, and the difference is that the ``{Trans.}'' model is first pre-trained on sentence-level NMT corpus and then is fine-tuned on the chat translation data.

\paragraph{Context-aware NMT Systems.} 

{Dia-Trans.}\\~\cite{maruf-etal-2018-contextual}: A Transformer-based model where an additional encoder is used to introduce the mixed-language dialogue context, re-implemented by~\citet{liang-etal-2021-modeling}.

{Gate-Trans.}~\cite{zhang-etal-2018-improving} and {NCT}~\cite{ma-etal-2020-simple}: Both are document-level NMT Transformer models where they introduce the dialogue history by a gate and by sharing the first encoder layer, respectively.

{CPCC}~\cite{liang-etal-2021-modeling}: A variational model that focuses on incorporating dialogue characteristics into a translator for better performance.

{CSA-NCT}~\cite{liang-etal-2021-towards}: A multi-task learning model that uses several auxiliary tasks to help generate dialogue-related translations.

\begin{table}[t!]
\centering
\newcommand{\tabincell}[2]{\begin{tabular}{@{}#1@{}}#2\end{tabular}}
\scalebox{0.80}{
\setlength{\tabcolsep}{0.9mm}{
\begin{tabular}{clllll}
\toprule
&\multicolumn{1}{c}{\multirow{2}{*}{\textbf{Models} (\emph{Base})}} &\multicolumn{2}{c}{$\textbf{En$\rightarrow$Zh}$}  &  \multicolumn{2}{c}{$\textbf{Zh$\rightarrow$En}$}     \\ 
\cmidrule(lr){3-4} \cmidrule(lr){5-6} 
&\multicolumn{1}{c}{} & \multicolumn{1}{c}{BLEU$\uparrow$} & \multicolumn{1}{c}{TER$\downarrow$} & \multicolumn{1}{c}{BLEU$\uparrow$} &  \multicolumn{1}{c}{TER$\downarrow$}  \\
\midrule
\multirow{5}{*}{\tabincell{c}{{Two-stage}\\\emph{w/o} data}}
&{Trans. \emph{w/o} FT} &21.40     &72.4 & 18.52    & 59.1          \\
&{Trans.}   &25.22 &62.8 & {21.59} &{56.7} \\\cdashline{2-6}[4pt/2pt]%\cline{1-10}
&{NCT}  & 24.76     &63.4  & 20.61   & 59.8  \\ 
&{M-NCT}   &\underline{27.84}      &\underline{59.8}   & \underline{22.41}    & \underline{55.9}       \\%\cdashline{2-6}[4pt/2pt]%\cline{1-10}
&{SML} (Ours)  &\textbf{28.96}$^{\dagger\dagger}$     &\textbf{58.3}$^{\dagger\dagger}$  & \textbf{23.23}$^{\dagger\dagger}$   & \textbf{55.2}$^{\dagger\dagger}$   \\
\midrule
\multirow{5}{*}{\tabincell{c}{{Three-stage}\\\emph{w/} data}}
&{Trans. \emph{w/o} FT}    &28.60  &56.7     &22.46  &53.9   \\
&{Trans.}      &30.90 &56.5 &25.04 &53.3 \\\cdashline{2-6}[4pt/2pt]%\cline{1-10}
&{NCT}   &31.37  &55.9   &25.35    &52.7  \\ 
&{M-NCT}   &\underline{31.63}   &\underline{55.6}   &\underline{25.86}   &\underline{51.9} \\ %\cdashline{2-6}[4pt/2pt]%\cline{1-10}
&{SML} (Ours)  &\textbf{32.25}$^{\dagger\dagger}$     &\textbf{55.1}$^{\dagger\dagger}$  & \textbf{26.42}$^{\dagger}$   & \textbf{51.4}$^{\dagger\dagger}$  \\ 
\bottomrule
\end{tabular}}}
\caption{Results on test sets of BMELD in terms of BLEU (\%) and TER (\%), where ``Two-stage \emph{w/o} data'' means the pre-training-then-fine-tuning paradigm and the in-domain data not being used, and ``Three-stage \emph{w/} data'' means the proposed three-stage method and this group uses the in-domain data. The ``M-NCT'' denotes the multi-task learning model jointly trained with four auxiliary tasks in a conventional manner. All models apply the same two/three-stage training strategy with our SML model for fair comparison except the ``Trans. \emph{w/o} FT'' model, respectively.}
\label{tbl:w_f}%\vspace{-10pt}
\end{table}

\begin{table}[t]
\centering
\scalebox{0.72}{
\begin{tabular}{llllll}
\toprule
\multirow{2}{*}{\#}& \multirow{2}*{\textbf{Where to Use?}}   & \multicolumn{2}{c}{\textbf{En$\rightarrow$Zh}} & \multicolumn{2}{c}{\textbf{Zh$\rightarrow$En}}  \\
\cmidrule(lr){3-4} \cmidrule(lr){5-6} 
& & BLEU$\uparrow$ & TER$\downarrow$ & BLEU$\uparrow$ & TER$\downarrow$\\
\midrule
0&Two-stage (Not Use) & 29.49  & 55.8  & 24.15 & 53.3 \\
1&Two-stage ({\textcircled{\small{1}}}) & 31.17  & 53.2  & 26.14 & 51.4 \\
2&Two-stage ({\textcircled{\small{2}}}) &29.87  & 53.7  & 27.47 & 50.5     \\\cdashline{2-6}[4pt/2pt]
3&Three-stage ({\textcircled{\small{2}}}) & \textbf{33.45}$^{\dagger\dagger}$  & \textbf{51.1}$^{\dagger\dagger}$ &\textbf{29.47}$^{\dagger\dagger}$ & \textbf{49.3}$^{\dagger\dagger}$\\
\bottomrule
\end{tabular}}
\caption{Results on validation sets of where to use the large-scale in-domain data under the \emph{Base} setting. The rows 0$\sim$2 use the pre-training-then-fine-tuning (\emph{i.e.}, two-stage) paradigm while row 3 is the proposed three-stage method. For a fair comparison, the final fine-tuning stage of rows 0$\sim$3 is all trained in the conventional multi-task training manner and the only difference is the usage of the in-domain data. Specifically, row 0 denotes without using the in-domain data. Row 1 denotes that we incorporate the in-domain data into the first pre-training stage ({\textcircled{\small{1}}}). Row 2 denotes that we introduce the in-domain data into the fine-tuning stage ({\textcircled{\small{2}}}). Row 3 denotes that we add a second pre-training stage to introduce the in-domain data.}
\label{tbl:data_usage}%\vspace{-10pt}
\end{table}

\subsection{Main Results}
\autoref{tbl:main_res} shows the main results on En$\leftrightarrow$Zh and En$\leftrightarrow$De under \emph{Base} and \emph{Big} settings. In~\autoref{tbl:w_f}, we present additional results on En$\leftrightarrow$Zh.%For comparison, as in \autoref{ssec:layout}, ``Trans. \emph{w/o} FT'' and ``Trans.'' are sentence-level baselines while ``Dia-Trans.'', ``Gate-Trans.'', ``NCT'', ``CPCC'', and ``CSA-NCT'' are the existing context-aware NMT models. Particularly, ``SML'' represents our proposed approach.

%%%%%%%%%%%%%%%%%%%%%%%%%%%%%%%%%%%%%%%%%%%%%%%%% Zh-En %%%%%%%%%%%%%%%%%%%%%%%%%%%%%%%%%%%%%%%%%%%%%%%%
\noindent\textbf{Results on En$\leftrightarrow$Zh.}
\label{ssec:ende}
Under the \emph{Base} setting, our SML significantly surpasses the sentence-level/context-aware baselines (\emph{e.g.}, the existing best model ``CSA-NCT''), 4.58$\uparrow$ on En$\rightarrow$Zh and 4.06$\uparrow$ on Zh$\rightarrow$En, showing the effectiveness of the large-scale in-domain data and our scheduled multi-task learning. In terms of TER, the SML also performs best on the two directions, 5.0$\downarrow$ and 4.3$\downarrow$ than ``CPCC'' (the lower the better), respectively. Under the \emph{Big} setting, the SML model consistently outperforms all previous models once again. 

%%%%%%%%%%%%%%%%%%%%%%%%%%%%%%%%%%%%%%%%%%%%%%%%% En-De %%%%%%%%%%%%%%%%%%%%%%%%%%%%%%%%%%%%%%%%%%%%%%%%

\paragraph{Results on En$\leftrightarrow$De.}
\label{ssec:chen}
On both En$\rightarrow$De and De$\rightarrow$En under the \emph{Base} setting, the SML approach presents remarkable improvements over other existing comparison methods by up to 2.50$\uparrow$ and 2.69$\uparrow$ BLEU gains, and by 2.55$\uparrow$ and 2.53$\uparrow$ BLEU gains under the \emph{Big} setting, respectively. This shows the superiority of our three-stage training framework and also demonstrate the generalizability of the proposed approach across different language pairs. Since the baselines of En$\leftrightarrow$De are very strong, the results of En$\leftrightarrow$De are not so significant than En$\leftrightarrow$Zh.

\paragraph{Additional Results.}
~\autoref{tbl:main_res} presents our overall model performance, though, strictly speaking, it is unfair to directly compare our approaches with previous ones. Therefore, we conduct additional experiments in~\autoref{tbl:w_f} under two settings: (\emph{\romannumeral1}) using the original pre-training-then-fine-tuning framework without introducing the large-scale in-domain data (\emph{i.e.}, ``Two-stage \emph{w/o} data'' group); (\emph{\romannumeral2}) using the proposed three-stage method with the large-scale in-domain data (\emph{i.e.}, ``Three-stage \emph{w/} data'' group). And we conclude that (1) the same model (\emph{e.g.}, SML) can be significantly enhanced by the second in-domain pre-training stage, demonstrating the effectiveness of the second pre-training on the in-domain data; (2) our SML model always exceeds the conventional multi-task learning model ``M-NCT'' in both settings, indicating the superiority of the scheduled multi-task learning strategy.
\section{Analysis}

\subsection{Ablation Study}
We conduct ablation studies in~\autoref{tbl:data_usage} and~\autoref{tbl:t_manner} to answer the following two questions. \textbf{Q1}: \textit{\textbf{why a three-stage training framework?}} and \textbf{Q2}: \textit{\textbf{why the scheduled multi-task learning strategy?}}

To answer \textbf{Q1}, in~\autoref{tbl:data_usage}, we firstly investigate the effect of the large-scale in-domain chat translation data and further explore where to use it. Firstly, the results of rows 1$\sim$3 substantially outperform those in row 0, proving the availability of incorporating the in-domain data. Secondly, the results of row 3 significantly surpass rows 1$\sim$2, indicating that the in-domain data used in the proposed second stage of our three-stage training framework is very successful rather than used in the stage of pre-training-then-fine-tuning paradigm. That is, the experiments show the effectiveness and necessity of our three-stage training framework.

To answer \textbf{Q2}, we investigate multiple multi-task learning strategies in~\autoref{tbl:t_manner}. Firstly, the results of row 3 are notably higher than those of rows 0$\sim$2 in both language directions, obtaining significant cumulative benefits of auxiliary tasks than rows 0$\sim$2, demonstrating the validity of the proposed SML strategy. Secondly, the results of row 3 vs row 4 show that the inverse gradient projection of auxiliary tasks also has a positive impact on the model performance, which may prevent the model from overfitting, working as a regularizer. All experiments show the superiority of our scheduled multi-task learning strategy.

\begin{table}[t!]
\centering
\scalebox{0.66}{
\setlength{\tabcolsep}{0.90mm}{
\begin{tabular}{llllll}
\toprule
\multirow{2}{*}{\#}&{\multirow{2}*{\quad\quad\quad \textbf{Training Manners?}}} &\multicolumn{2}{c}{$\textbf{En$\rightarrow$Zh}$}  &  \multicolumn{2}{c}{$\textbf{Zh$\rightarrow$En}$}    \\ 
\cmidrule(lr){3-4} \cmidrule(lr){5-6} 
&\multicolumn{1}{c}{} & \multicolumn{1}{c}{BLEU$\uparrow$} & \multicolumn{1}{c}{TER$\downarrow$} & \multicolumn{1}{c}{BLEU$\uparrow$} & \multicolumn{1}{c}{TER$\downarrow$}   \\ 
\midrule
% 60.81	24.8	62.32	24.9
0&{Conventional Multi-task Learning}   & {33.45}  & {51.2} &{29.47}  & {49.3} \\ %$\pm$0.31 
1&Random Multi-task Learning & 32.88  & 51.6  & 29.19 & 49.5      \\
2&Prior-based Multi-task Learning & 33.94  & 51.1  & 29.74 & 49.1     \\\cdashline{2-6}[4pt/2pt]
% \midrule
3& Scheduled Multi-task Learning (SML)  &\textbf{34.21}$^{\dagger}$  & \textbf{51.0}   & \textbf{30.13}$^{\dagger}$&\textbf{49.0}\\ 
4& SML \emph{w/o} inverse gradient projection &33.85  &51.1  & 29.79 & 49.1     \\ 
\bottomrule
\end{tabular}}}
\caption{Results on validation sets of the three-stage training framework in different multi-task training manners, under the \emph{Base} setting. Row 1 denotes that the auxiliary tasks are randomly added in a conventional training manner at each training step. Row 2 denotes that we add the auxiliary tasks according to their performance in different stages, \emph{i.e.}, we add all tasks in the second stage while only considering MRG and NUD in the fine-tuning stage according to prior trial results in~\
\autoref{fig.zx}. Row 4 denotes that we remove the inverse gradient projection of auxiliary tasks (\emph{i.e.},~\autoref{fig:gradient} (b)).}
\label{tbl:t_manner}%\vspace{-10pt}
\end{table}

%%%%%%%%%%%%%%%%%%%%%%%%%%%%%%%%%%%%%%%%%%%%%%%%% Human %%%%%%%%%%%%%%%%%%%%%%%%%%%%%%%%%%%%%%%%%%%%%%%%
\subsection{Human Evaluation}
\label{ssec:he}
Inspired by previous work~\cite{bao-EtAl:2020:WMT,liang-etal-2021-modeling}, we apply two criteria for human evaluation to judge whether the translation result is: 
\begin{enumerate}[itemindent=1em]
\item semantically \textbf{coherent} with the dialogue context? 
% \item preserves the \textbf{domain} property? 
\item grammatically correct and \textbf{fluent}? 
\end{enumerate}

Firstly, we randomly sample 200 conversations from the test set of BMELD in En$\rightarrow$Zh. Then, we use 6 framework in \autoref{human_evaluation} to generate translated utterances of these sampled conversations. Finally, we assign the translated utterances and their corresponding dialogue context in the target language to three postgraduate student annotators, and then ask them to make evaluations (0/1 score) according to the above two criteria, and average the scores as the final result.

\autoref{human_evaluation} shows that the SML produces more coherent and fluent translations than other comparison models (significance test, {\em p} \textless \ 0.05), which shows the effectiveness of our proposed method. The inter-annotator agreements are  0.558 and 0.583 for {coherence} and {fluency} calculated by the Fleiss’ kappa~\cite{doi:10.1177/001316447303300309}, respectively. It indicates ``Moderate Agreement'' for both criteria.

% Besides, we investigate the automatic coherence evaluation in~\autoref{Dia_coherence}, which also demonstrates that our approach can produce evidently more coherent translations than comparison models.
\begin{table}[t]
\centering
\newcommand{\tabincell}[2]{\begin{tabular}{@{}#1@{}}#2\end{tabular}}
% \small
\scalebox{0.80}{
\setlength{\tabcolsep}{4.0mm}{
\begin{tabular}{lll}
\toprule
\multirow{1}{*}{\textbf{Models} (\emph{Base}}) & \multicolumn{1}{c}{$\textbf{Coherence}$} &  \multicolumn{1}{c}{$\textbf{Fluency}$} \\%\cline{1-4}
\midrule
Trans. \emph{w/o} FT        &0.585  &0.630 \\
Trans.     &0.620  &0.655 \\\cdashline{1-3}[4pt/2pt]
NCT &0.635   &0.665 \\
CSA-NCT     &{0.650} &{0.680} \\%\cdashline{2-4}[4pt/2pt]
M-NCT     &{0.665} &{0.695} \\%\cdashline{2-4}[4pt/2pt]
SML (Ours)       &\textbf{0.690}$^{\dagger}$ &\textbf{0.735}$^{\dagger}$ \\
\bottomrule
\end{tabular}}}
\caption{Results of human evaluation ({En$\rightarrow$Zh}). All models use the three-stage training framework to introduce the in-domain data.}
\label{human_evaluation} %\vspace{-15pt}
\end{table}
\label{ssec:cs}

%%%%%%%%%%%%%%%%%%%%%%%%%%%%%%%%%%%%%%%%%%%%%%%%% WMT %%%%%%%%%%%%%%%%%%%%%%%%%%%%%%%%%%%%%%%%%%%%%%%%
\label{ssec:dc}
\begin{table}[t]
\centering
\newcommand{\tabincell}[2]{\begin{tabular}{@{}#1@{}}#2\end{tabular}}
% \small
% \scalebox{0.92}{
\setlength{\tabcolsep}{1.6mm}{
\begin{tabular}{llll}
\toprule
\multirow{1}{*}{\textbf{Models} (\emph{Base})} &  \multicolumn{1}{c}{\textbf{1-th Pr.}} & \multicolumn{1}{c}{\textbf{2-th Pr.}}  & \multicolumn{1}{c}{\textbf{3-th Pr.}}\\
\midrule
% V-Transformer   &-  &- &-\\
Trans. \emph{w/o} FT              &58.11  &55.15 &52.15\\
Trans.            &58.77  &56.10 &52.71  \\\cdashline{1-4}[4pt/2pt]
NCT        &59.19  &56.43 &52.89\\
CSA-NCT  &59.45  &{56.74} &{53.02}\\%\cdashline{2-4}[4pt/2pt]
M-NCT  &59.57  &{56.79} &{53.18}\\%\cdashline{2-4}[4pt/2pt]
SML (Ours)  &60.48$^{\dagger\dagger}$  &{57.88}$^{\dagger\dagger}$ &{53.95}$^{\dagger\dagger}$  \\\cdashline{1-4}[4pt/2pt]
% \midrule
Human Reference         &\textbf{61.03}  &\textbf{59.24} &\textbf{54.19}\\
\bottomrule
\end{tabular}}%}
\caption{Results (\%) of sentence similarity as dialogue coherence on validation set of BMELD in En$\rightarrow$Zh direction. All models use the three-stage training framework to introduce the in-domain data. The ``\#\textbf{-th Pr.}'' indicates the \#-th preceding utterance to the current one. ``$^{\dagger\dagger}$'' denotes that the improvement over the best result of other comparison models is statistically significant ({\em p} \textless \ 0.01). }
\label{coherence} %\vspace{-5pt}
\end{table}

\subsection{Dialogue Coherence}
\label{Dia_coherence}
We measure dialogue coherence as sentence similarity following~\citet{lapata2005automatic,Xiong_He_Wu_Wang_2019,liang-etal-2021-modeling}: 
\begin{equation}\nonumber
\begin{split}
% \label{eq:nmt}
    coh(s_1, s_2) &= \mathrm{cos}(f({s_1}), f({s_2})),%\\ 
    % f(s_i) &= \frac{1}{\vert f(s_i)\vert}\sum_{\vec{w} \in s_i}(\vec{w}),
\end{split}
\end{equation}
where $\mathrm{cos}$ denotes cosine similarity and $f(s_i) = \frac{1}{\vert s_i\vert}\sum_{\textbf{w} \in s_i}(\textbf{w})$ and \(\textbf{w}\) is the vector for word $w$, and $s_i$ is the sentence. Then, the Word2Vec\footnote{https://code.google.com/archive/p/word2vec/}~\cite{mikolov2013efficient} is applied to obtain the distributed word vectors (dimension size is 100), which is trained on our conversation dataset\footnote{We choose our constructed dialogue corpus to learn the word embedding.}.

\autoref{coherence} presents the results of different models on validation set of BMELD in En$\rightarrow$Zh direction in terms of coherence. It demonstrates that the proposed SML model generate more coherent translations compared to other previous models (significance test, {\em p} \textless \ 0.01).

\subsection{Effect of the Auxiliary Task: XNUD}
\label{XNUD}
We investigate the effect of the XNUD task. As shown in~\autoref{tbl:xnud}, the ``M-NCT'' denotes the multi-task learning model jointly trained with four auxiliary tasks in conventional manner.  After removing the XNUD task, the performance drops to some extend, indicating that the new XNUD task achieves further performance improvement based on three existing auxiliary tasks~\cite{liang-etal-2021-towards}. Then, based on the strong ``M-NCT'' model, we further investigate where and how to make the most of them for the main NCT task.

\begin{table}[t]
\centering
\scalebox{0.65}{
\begin{tabular}{lllll}
\toprule
\multirow{2}*{\textbf{Models} (\emph{Base})}   & \multicolumn{2}{c}{\textbf{En$\rightarrow$Zh}} & \multicolumn{2}{c}{\textbf{Zh$\rightarrow$En}}  \\
\cmidrule(lr){2-3} \cmidrule(lr){4-5} 
  & BLEU$\uparrow$ & TER$\downarrow$ & BLEU$\uparrow$ & TER$\downarrow$\\
\midrule
NCT+\{MRG,CRG,NUD\} & {28.94}  & {56.0} &{23.82}    & {54.3}  \\
NCT+\{MRG,CRG,NUD,XNUD\}&\bf{29.49}$^{\dagger\dagger}$  & \bf{55.8}  & \bf{24.15}$^{\dagger}$ & \bf{53.5}$^{\dagger\dagger}$\\
\bottomrule
\end{tabular}}
\caption{The results on validation sets after adding the XNUD task on three auxiliary tasks, \emph{i.e.}, MRG, XRG and NUD~\cite{liang-etal-2021-towards}, which are trained in conventional manner (without incorporating in-domain data).}
\label{tbl:xnud}
\end{table}

%%%%%%%%%%%%%%%%%%%%%%%%%%%%%%%%%%%%%%%%%%%%%%%%% Related Work %%%%%%%%%%%%%%%%%%%%%%%%%%%%%%%%%%%%%%%%%%%%%%%%
\section{Related Work}
\paragraph{Neural Chat Translation.} The goal of NCT is to train a dialogue-aware translation model using the bilingual dialogue history, which is different from document-level/sentence-level machine translation~\cite{maruf-etal-2019-selective,ma-etal-2020-simple,yanetal2020multi,meng2019dtmt,zhangetal2019bridging}. Previous work can be roughly divided into two categories. One~\cite{maruf-etal-2018-contextual,lrec,rikters-etal-2020-document,9023129} mainly pays attention to automatically constructing the bilingual corpus since no publicly available human-annotated data~\cite{farajian-etal-2020-findings}. The other~\cite{wang2021autocorrect,liang-etal-2021-modeling,liang-etal-2021-towards} aims to incorporate the bilingual dialogue characteristics into the NCT model via multi-task learning. Different from the above studies, we focus on introducing the in-domain chat translation data to learn domain-specific patterns and scheduling the auxiliary tasks to exert their potential for high translation quality. 

\paragraph{Multi-task Learning.} Conventional multi-task learning~\cite{Multitask}, which mainly focuses on training a model on multiple related tasks to promote the representation performance of the main task, has been successfully used in many natural language processing tasks~\cite{10.1145/1390156.1390177,DBLP:journals/corr/Ruder17a,Deng2013NewTO,liang2020infusing,liang-etal-2021-iterative-multi,liang2020dependency}. In the NCT, conventional multi-task learning has been explored to inject the dialogue characteristics into models with dialogue-related tasks such as response generation~\cite{liang-etal-2021-modeling,liang-etal-2021-towards}. In this work, we instead focus on how to schedule the auxiliary tasks at training to make the most of them for better translations. 

%%%%%%%%%%%%%%%%%%%%%%%%%%%%%%%%%%%%%%%%%%%%%%%%% Conclusions %%%%%%%%%%%%%%%%%%%%%%%%%%%%%%%%%%%%%%%%%%%%%%%%
\section{Conclusion}
This paper proposes a scheduled multi-task learning framework armed with an additional in-domain pre-training stage and a gradient-based scheduled multi-task learning strategy. Experiments on En$\leftrightarrow$Zh and En$\leftrightarrow$De demonstrate that our framework significantly improves translation quality in terms of BLEU and TER metrics, proving its effectiveness and generalizability. Human evaluation also proves that the proposed approach yields better translations in terms of coherence and fluency. Furthermore, we contribute two large-scale in-domain paired bilingual dialogue datasets to the research community.

\section*{Acknowledgements}
This work is supported by the National Key R\&D Program of China (2020AAA0108001) and the National Nature Science Foundation of China (No. 61976015, 61976016, 61876198 and  61370130). Yunlong Liang is supported by 2021 Tencent Rhino-Bird Research Elite Training Program. The authors would like to thank the anonymous reviewers for their insightful comments and suggestions to improve this paper.

% Entries for the entire Anthology, followed by custom entries
% better translation in terms of coherence and fluency.
\bibliography{anthology,custom}
\bibliographystyle{acl_natbib}

% \clearpage
% \newpage
\appendix
\label{sec:appendix}

\section{Datasets}
\label{Appendix_dataset}
As pointed in \autoref{sect:data}, our training datasets involve the WMT20 dataset for general-domain pre-training, the newly constructed in-domain chat translation data for the second pre-training (please refer to~\autoref{sec:ts}), and two target chat translation corpora, BMELD~\cite{liang-etal-2021-modeling} and BConTrasT~\cite{farajian-etal-2020-findings}. The statistics of the splits of training, validation, and test sets of BMELD (En$\leftrightarrow$Zh) and BConTrasT (En$\leftrightarrow$De) are shown in \autoref{datasets}.

\paragraph{WMT20.} Following previous work~\cite{liang-etal-2021-modeling,liang-etal-2021-towards}, for En$\leftrightarrow$Zh, we combine News Commentary v15, Wiki Titles v2, UN Parallel Corpus V1.0, CCMT Corpus, and WikiMatrix. For En$\leftrightarrow$De, we combine six corpora including Euporal, ParaCrawl, CommonCrawl, TildeRapid, NewsCommentary, and WikiMatrix. Firstly, we filter out duplicate sentence pairs and remove those whose length exceeds 80. To pre-process the raw data, we employ a series of open-source/in-house scripts, including full-/half-width conversion, unicode conversation, punctuation normalization, and tokenization~\cite{wang-EtAl:2020:WMT1}. After filtering, we utilize BPE~\cite{sennrich-etal-2016-neural} with 32K merge operations. Finally, we get 22,244,006 sentence pairs for En$\leftrightarrow$Zh and 45,541,367 sentence pairs for En$\leftrightarrow$De, respectively. 

\paragraph{BMELD.}~\citet{liang-etal-2021-modeling} construct this English$\leftrightarrow$Chinese bilingual dialogue dataset. Specifically, based on the dialogue dataset in the MELD (originally in English)~\cite{poria-etal-2019-meld}, they firstly crawled\footnote{\url{https://www.zimutiantang.com/}} the corresponding Chinese translations and then manually post-edited them according to the dialogue context by native Chinese speakers (post-graduate students majoring in English). Finally, they follow the usage of BConTrasT~\citet{farajian-etal-2020-findings} and assume 50\% speakers as Chinese speakers to keep data balance for Zh$\rightarrow$En translations and thus build the \underline{b}ilingual MELD (BMELD).

\paragraph{BConTrasT.} The BConTrasT dataset\footnote{https://github.com/Unbabel/BConTrasT} is first provided by WMT 2020 Chat Translation Task~\cite{farajian-etal-2020-findings}, based on one monolingual Taskmaster-1 corpus~\cite{byrne-etal-2019-taskmaster}, which is  first automatically translated into German and then manually post-edited by Unbabel editors\footnote{www.unbabel.com} who are native German speakers. Then, having the conversations in two languages allows us to simulate bilingual conversations in which one speaker (agent), speaks in English and the other speaker (customer), responds in German.

%%%%%%%%%%%%%%%%%%%%%%%%%%%%%% Datasets %%%%%%%%%%%%%%%%%%%%%%%%%%%%%%%%%
\begin{table}[t]
\centering
\newcommand{\tabincell}[2]{\begin{tabular}{@{}#1@{}}#2\end{tabular}}
\small
% \scalebox{0.9}{
\setlength{\tabcolsep}{1.8mm}{
\begin{tabular}{lrrrrrr}
\toprule
\multirow{2}{*}{\bf{Datasets}} & \multicolumn{3}{c}{\#\bf{Dialogues}} &  \multicolumn{3}{c}{\#\bf{Utterances}} \\
\cmidrule(lr){2-4} \cmidrule(lr){5-7}
&Train &Valid & Test&Train &Valid & Test\\\hline
En$\rightarrow$Zh    &1,036&108&274  &5,560&567&1,466 \\
Zh$\rightarrow$En    &1,036&108&274   &4,427&517&1,135 \\
En$\rightarrow$De    &550&78&78 &7,629 &1,040 &1,133\\
De$\rightarrow$En    &550&78&78  &6,216 &862 &967\\
\bottomrule
\end{tabular}}%}
\caption{Statistics of the chat translation data.
}\label{datasets}
\end{table}

\textbf{\begin{figure*}[t]
    \centering
    \includegraphics[width=1.0\textwidth]{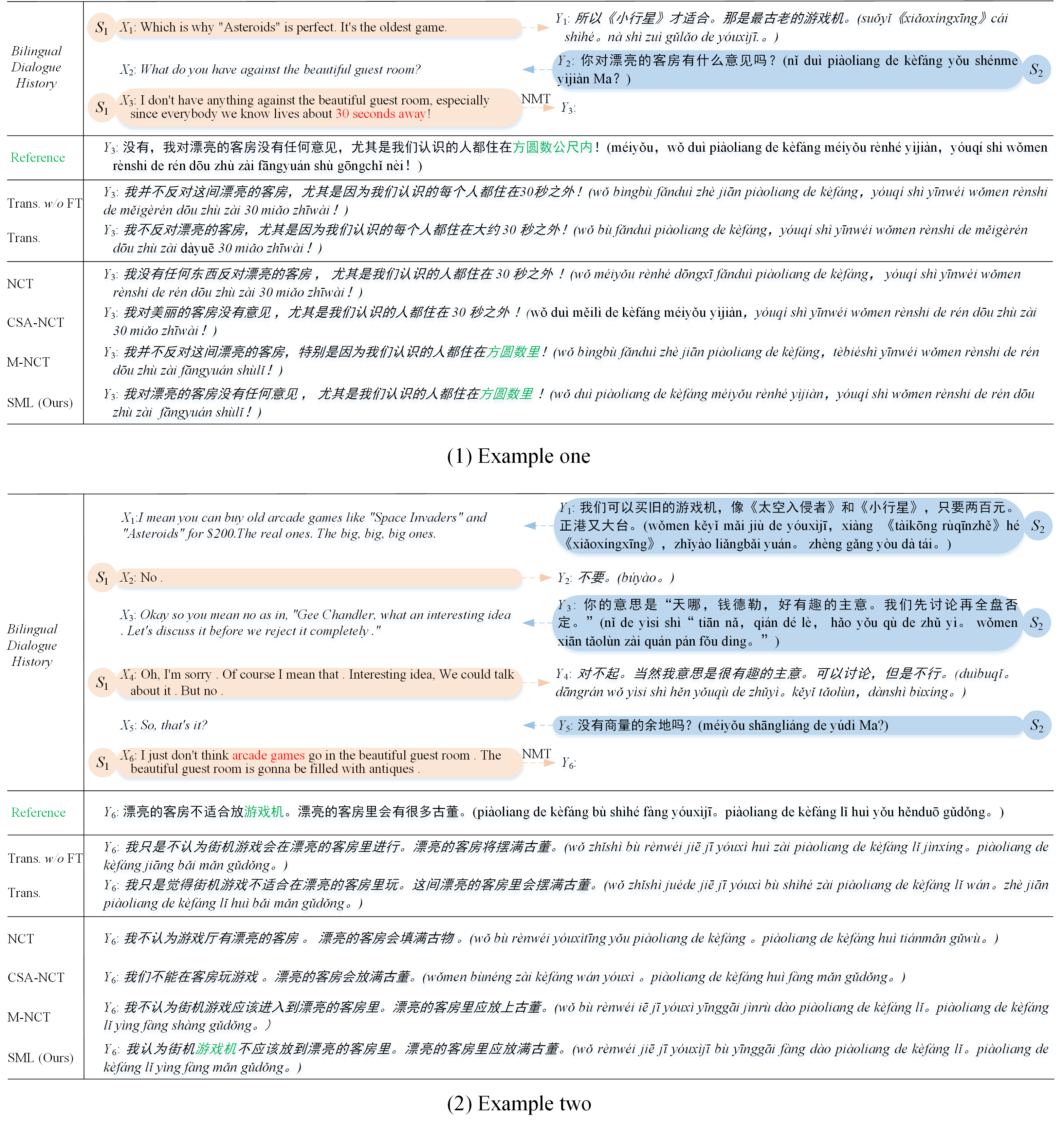}
    \caption{The illustrative cases of bilingual conversation translation.
    }
    \label{fig.case}%\vspace{-10pt}
\end{figure*}}

\section{Implementation Details}
\label{ID}
For all experiments, we follow the settings of~\citet{vaswani2017attention}, namely \emph{Transformer-Base} and \emph{Transformer-Big}. 
In \emph{Base} setting, we utilize 512 as hidden size (\emph{i.e.}, $d$), 2048 as filter size,  and 8 heads in multihead attention. In \emph{Big} setting, we use 1024 as hidden size, 4096 as filter size, and 16 heads in multihead attention. All Transformer models in this paper include $L$ = 6 encoder layers and the same decoder layers and all models are implemented with THUMT~\cite{tan-etal-2020-thumt} framework. For a fair comparison, we set the training step for the first pre-training stage and the second pre-training stage totally to 200,000 (100,000 for each stage), and set the step of fine-tuning stage 5,000. As for the balancing factor $\alpha$ in \autoref{loss_new}, we follow~\cite{liang-etal-2021-towards} to decay $\alpha$ from 1 to 0 over training steps (we set them to 100,000 and 5,000 for the last two training stages, respectively). The batch size on each GPU is 4096 tokens. All experiments in three stages are conducted utilizing 8 NVIDIA Tesla V100 GPUs, which gives us about 8*4096 tokens per update for all experiments. We use Adam~\cite{kingma2017adam} with $\beta_1$ = 0.9 and $\beta_2$ = 0.998 for all models, and set learning rate to 1.0, and set label smoothing to 0.1. We set dropout to 0.1/0.3 for \emph{Base} and \emph{Big} setting, respectively. When building the shared vocabulary $|V|$, we keep such word if its frequency is larger than 100. $|T|$ is set to 10.
The BLEU score on validation sets is selected as criterion for searching hyper-parameter. At inference, the beam size is set to 4, and the length penalty is 0.6 among all experiments. 

In the case of blind testing or online use (assumed dealing with En$\rightarrow$De), since translations of target utterances (\emph{i.e.}, English) will not be given, an inverse De$\rightarrow$En model is simultaneously trained and used to back-translate target utterances~\cite{bao-EtAl:2020:WMT}, which is similar for other translation directions.

\section{Case Study}
In this section, we present two illustrative cases in~\autoref{fig.case} to give some observations among the comparison models and ours.

For the case~\autoref{fig.case} (1), we find that most comparison models just translate the phrase ``30 seconds away'' literally as ``30 秒之外 (30 \emph{mi\v{a}o zh\={i}w\`{a}i})'', which is very strange and is not in line with Chinese language habits. By contrast, the ``M-NCT'' and ``SML'' models, through three-stage training, capture such translation pattern and generate an appropriate Chinese phrase ``方圆数里 (f\={a}ngy\'uan sh\`ul\v{i})''. The reason behind this is that the large-scale in-domain dialogue bilingual corpus contains many cases of free translation, which is common in daily conversations translation. This suggests that the in-domain pre-training is indispensable for a successful chat translator.

For the case~\autoref{fig.case} (2), we observe that the comparison models fail to translate the word ``games'', where they translate it as ``游戏 (y\'{o}ux\`{i})''. The reason may be that they cannot fully understand the dialogue context even though some models (\emph{e.g.}, ``CSA-NCT'' and ``M-NCT'') also jointly trained with the dialogue-related auxiliary tasks. By contrast, the ``SML'' model, enhanced by multi-stage scheduled multi-task learning, obtains accurate results. 

In summary, the two cases show that our SML model enhanced by the in-domain data and scheduled multi-task learning yields satisfactory translations, showing its superiority.

\quad
\quad
\quad
\end{CJK}
\end{document}